\documentclass[pdflatex,sn-mathphys-num]{sn-jnl}

\usepackage{graphicx}
\usepackage{multirow}
\usepackage{amsmath,amssymb,amsfonts}
\usepackage{mathrsfs}
\usepackage[title]{appendix}
\usepackage{xcolor}
\usepackage{textcomp}
\usepackage{manyfoot}
\usepackage{booktabs}
\usepackage{pdflscape}

\begin{document}

\title[Article Title]{Benchmarking the Connectomes of Caenorhabditis elegans within the Reservoir Computing Framework}

\author[1]{\fnm{Felix} \spfx{S.} \sur{Reimers}}\email{felix.s.reimers@hiof.no}

\author[]{\fnm{Ola} \spfx{Huse} \sur{Ramstad}}\email{ola.h.ramstad@gmail.com}

\author[2,3]{\fnm{Aliaksandr} \sur{Hubin}}\email{aliaksandr.hubin@nmbu.no}

\author*[1]{\fnm{Stefano} \sur{Nichele}}\email{stefano.nichele@hiof.no}

\affil*[1]{\orgdiv{Department of Computer Science and Communication}, \orgname{Østfold University College}, \orgaddress{\city{Halden}, \postcode{1757}, \state{Østfold}, \country{Norway}}}

\affil[2]{\orgdiv{Faculty of Chemistry, Biotechnology and Food Science}, \orgname{Norwegian University of Life Sciences}, \orgaddress{\city{Ås}, \postcode{1432}, \state{Akershus}, \country{Norway}}}

\affil[3]{\orgdiv{Department of Mathematics}, \orgname{University of Oslo}, \orgaddress{\city{Oslo}, \postcode{0316}, \state{Oslo}, \country{Norway}}}

\abstract{
The aim of this work is to examine the connectomes of Caenorhabditis elegans through a computational lens using the reservoir computing framework. Connectomes are mappings of biological neural networks; C. elegans is the first organism for which physical connectomes covering the whole nervous system have been published. The connectomes of C. elegans used in this paper have been derived at different ages of the organism and are based on three different ways of measuring inter-cellular connections. They have, with minimal preprocessing, been implemented as reservoirs in the form of echo state networks, which are recurrent neural networks. In reservoir computing, the reservoir itself is not trained, rather the output of the reservoir is passed to a comparatively small read-out module in which training takes place. Training and testing is conducted in different neuro-inspired tasks, with the aim of using these tasks as a benchmark for the connectomes.
This process has been repeated with different configurations of the reservoir and equally sized but randomized null models have been used for comparison.
The results show that the biological wiring and a bio-informed configuration of input and output nodes of the reservoirs do not necessarily lead to better performance. Contrarily, the randomized null models are often outperforming the original connectomes on the chosen benchmarks. At the same time it becomes clear that the results depend a lot on the configuration of the reservoir and the way the connectome has been derived from the organism. Connectomes from different ages may produce varying outcome, without a clear trend becoming visible.}

\keywords{reservoir computing, echo state network, machine learning, connectomes, Caenorhabditis elegans, neuro-computation}

\maketitle

\section{Introduction}\label{sec:intro}

The nematode Caenorhabditis elegans has a neural system whose physical structure has been completely mapped, resulting in the first full connectome of an animal. Such connectomes can be interpreted as the topology of a network, including a labeling of the nodes according to their biological function. \cite{white1986structure}

While these networks can be analyzed with graph theoretical measures, this paper takes a different approach: The connectomes are treated as a computational substrate and it is tested whether a benchmark can be developed that connects their computational performance on an array of tasks to the state of the network.

For this, the framework of reservoir computing is used. The development of the reservoir computing framework has been motivated by biological computation in the first place \cite{jaeger_echo_2001, maass_real-time_2002}. In it, untrained recurrent networks, among other systems, are used to project input data into a higher dimensional space on which a comparatively simple read-out module is trained. This makes this framework largely agnostic to the type of system used and only comparatively little resources are needed for training.

The research questions of this work are: Do the networks derived from the C. elegans connectome show beneficial computational performance when benchmarked within the reservoir computing framework? Does a biologically-informed wiring and choice of input and output nodes increase performance? These questions are motivated by the idea that a biological system formed by evolution should differ from equally sized random systems with regards to its computational performance. An additional question is whether connectomes derived at different ages show differences in performance. This aims at seeing if changes in the state of a network, such as aging, are reflected in the computational benchmark. In order to answer these questions, C. elegans connectomes are implemented as reservoirs and trained and tested in different neuro-inspired tasks. The performance is then compared with null models, which are randomized versions of the connectomes. The aim is not to train for good performance, but to see if a benchmark can be developed in this way that requires little preprocessing of the underlying networks. Benchmarking in-silico networks might yield information that can be used to better set up biological computers or serve as inspiration for the development of digital learning systems.

The rest of the work is structured as follows: In section \ref{sec:related}, some related work is presented. Next, the technical details of the implementation of this work are presented in section \ref{sec:framework}. Afterwards, results are presented and discussed in section \ref{sec:results}. Lastly, the results are summarized and an outlook for possible follow-up examination is presented in section \ref{sec:conclusion}. The supplementary material contains a link to the code repository.

\section{Related Work}\label{sec:related}

The question of whether the state of a network can be derived from its computational performance has been the target of previous work, where, in a proof-of-concept manner, perturbations of the mobile communication network of Norway have been examined with a very similar framework to the one presented in this work. Here, nodes were iteratively dropped from the network; these perturbations lead to a measurable decrease in performance \cite{reimers2025benchmarking}.

Fundamental on the technical side is the \textit{conn2res}, short for connectome-to-reservoir, toolbox \cite{suarez_connectome_2024}. It allows for easy implementation of connectomes as reservoirs while at the same time giving the user the flexibility to choose different hyperparameter and model architectures. As it is made for connectomes, it is possible to easily leverage information on the biological labeling of nodes. Furthermore, a number of tasks from the NeuroGym \cite{Crocioni_neurogym_2025} and ReservoirPy \cite{trouvain2020reservoirpy} package are natively supported. Details on the implementation of this work can be found in section \ref{sec:framework}.

Other previous work explores the usage of in-vitro grown cell cultures as reservoirs or even the use of cell cultures as an actively trained substrate \cite{lindell_information_2024, cai_brain_2023, kagan_vitro_2022}. These approaches rely on using living material as in-vitro computational substrate, whereas this works explores the computational performance of a digital snapshots of a connectome, which is furthermore not derived from a cell culture but an organism. The mentioned work inspired the computational perspective on connectomes as reservoirs. As growing cell cultures in a laboratory is a difficult and taxing endeavor, the comparatively simple work with in-silico networks might help in creating targets for the creation of biological computers and a simple benchmark could give an opportunity for early testing and intervention.

\section{Framework}\label{sec:framework}

In this section, the C. elegans connectomes and the way they are implemented as reservoirs is described in more detail. Afterwards, the details of how individual experiments were set up and analyzed is described.  The following Figure \ref{fig:design} gives an overview of the different design choices described below.

\begin{figure}[h]
\centering
\includegraphics[width=0.7\textwidth]{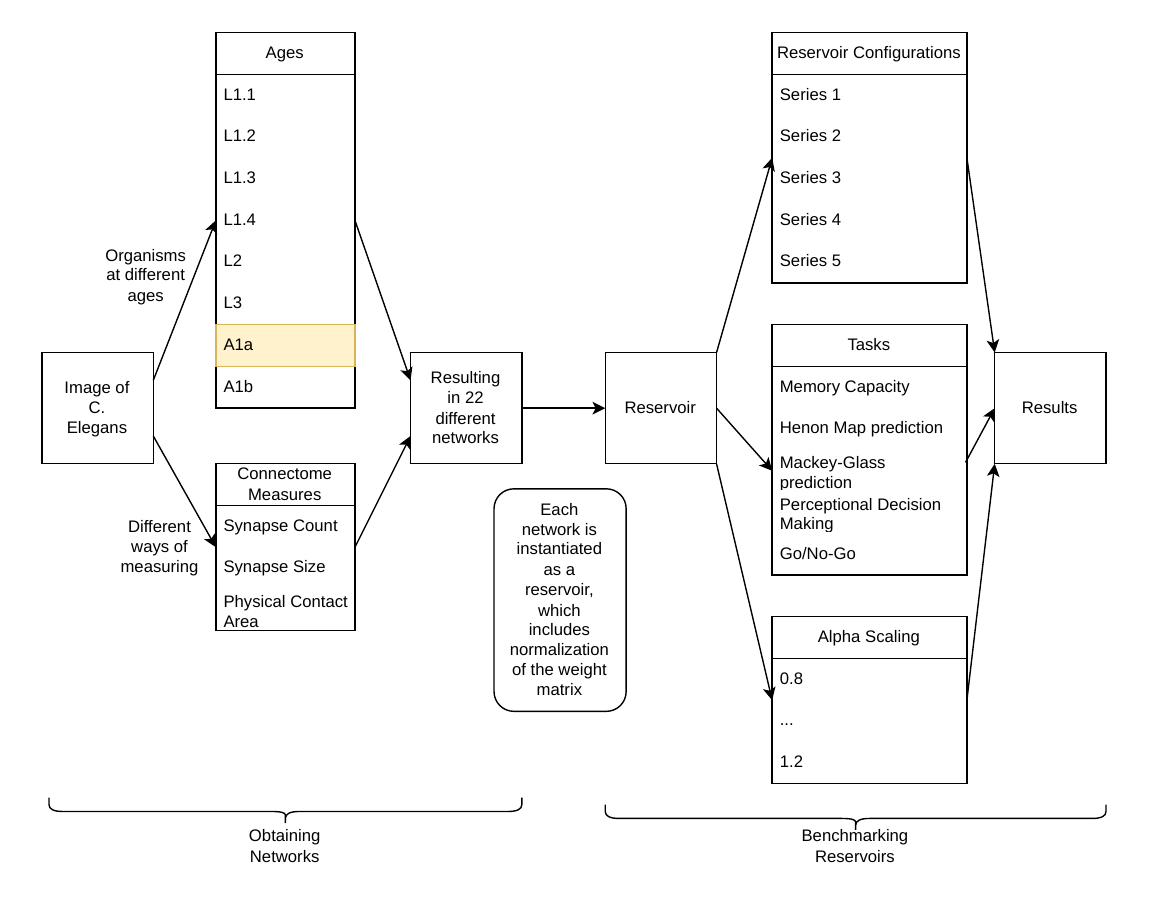}
\caption{The different hyperparameter and configuration choices that have been covered in the experiments. On the left of the figure it is described in which way the reservoirs are derived, on the right of the figure the conditions under which they are tested is depicted. A1a in Ages is color marked as here only the synapse count based connectome is available.} \label{fig:design}
\end{figure}

The C. elegans connectomes used in this work are published in \cite{witvliet2021connectomes}. The individual networks consist of 180 brain neurons plus body-wall muscles, glia and two canal-associated neurons. The nodes are labeled with individual names and grouped according to function. An adjacency matrix contains information of how the neurons are wired. Three different types of measurements have been used to derive these matrices.
The first measurement counts the number of connecting synapses while the second is based on the size of connecting synapses. As the direction of the synapses is considered, the resulting networks are directed too. The third way to derive the connectomes that has been used is to measure the physical contact area between cells, which results in bidirectional networks.
As such, the connectomes map the physical structure of the neural system of C. elegans organisms, not how information traverses the system (which would be a functional connectome). Connectomes have been constructed for "eight isogenic Caenorhabditis elegans individuals across postnatal stages"\cite{witvliet2021connectomes}, yielding four connectomes for organisms at the larval L1 state at different points in time during that state (L1.1 to L1.4), one for the L2 state, one for the L3 state and two for individual adult organisms. For the networks derived from synaptic size and physical contact, one of the adult stages is missing. This means that in total 22 connectomes are examined. The authors describe that synaptic connections often form where physical contact is already present, as such the edges of the synapse based connectecomes are largely present within the physical contact area based connectome. Their examination furthermore suggests a similarity between synapse size and synapse count based connectomes. They further describe an increase of synaptic connection with age, which increase the amount of forward connections and modularity in the network, while interneuron circuits on the other side remain stable. Further differences between isogenic individuals are noticed. The physical contact on the other side is stable throughout the development of individuals. \cite{witvliet2021connectomes}

\begin{figure}[h]
\centering
\includegraphics{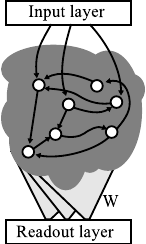}
\caption{Conceptual depiction of an echo state network. Only the matrix W, connecting the reservoir to the read-out layer, is trained. Taken from \cite{nichele_deep_2017}} \label{fig:ESN}
\end{figure}

The computational framework of reservoir computing goes back to independent work on two models, the echo state network \cite{jaeger_echo_2001} and the liquid state machine \cite{maass_real-time_2002}. In this work, the former model is used. Here, the network is modeled similarly to common recurrent artificial neural networks, with weight multiplications happening along the edges and non-linear functions being applied at the nodes. The strength of the physical connections between neurons and muscles is treated as the weights of the network. Before use, the adjacency matrix is normalized, such that the spectral radius of the matrix equals one. During the experiments, the weight matrices were scaled with an array of alpha values. In this work, the reported performance is the optimal value over all alpha values. During the training, the weights of this matrix are frozen and remain unchanged. An input matrix which also has fixed weights is used to project an input signal into chosen nodes of the reservoir. Downstream from the reservoir, a read-out module interprets the reservoirs state sampled from chosen output nodes. This is the only part where a learning happens. See Figure \ref{fig:ESN} for a conceptual overview of echo state networks. The readout module in this work is either a ridge regression or a classifier based on a ridge regression. Utilizing the conn2res toolbox, the input and output nodes of the reservoir can be picked according to their biological function with input neurons being picked among the sensory and output nodes among the body-wall muscles.

In order to answer the questions outlined in section \ref{sec:intro}, three different null models are created: For the first one, the connections are rewired such that the degree distribution of outgoing weights remains the same. The nodes are still picked from the same groups as for the original models. For the second null model, the original wiring is kept, but the nodes are picked randomly from all available nodes. In the third null model, both the edges of the network are rewired and the nodes are picked randomly. This means that for the second and third null model the input/output structure of the original connectomes is not considered, with the exception of the first reservoir configuration (see below). For every run of an original model, the null models are sampled 15 times, creating some random variation among them.
In the appendix, Figure \ref{fig:heatmap_orig} shows a heatmap of the connectome derived from the number of connecting synapses for the earliest larvae state while Figure \ref{fig:heatmap_rewired} shows it after rewiring.

Additionally, five different configurations of the reservoirs have been used. In the first series of tests, the number of input nodes corresponds to the number of input signals $x$, while the number of output nodes is arbitrarily fixed as five. Furthermore, only sensory nodes that are available for every age may be used, meaning that even for the second and third null model, only the output nodes are picked randomly. Series two is similar, but drops the requirement that input/output (I/O) nodes need to be available at every age. In series three, the number of I/O nodes is $50\%$ of the nodes belonging to the sensory or muscle group respectively, where the number of input nodes is rounded down to a multiple of the input signals of the tasks. For series four, a percentage of $80\%$ is chosen and for series five it is $30\%$. For the first two configurations, the input is presented to the chosen nodes without alteration. For series three to five, the weights of the input matrix are randomly sampled from a $[-1, 1]$-uniform distribution for the chosen input nodes during runtime.

Five different tasks with diverse demands to the learner are used as benchmarks. Most of them stem from the neurogym package, one is natively implemented in conn2res. The Memory Capacity task requires the model to repeat an input signal with a lag of up to twenty time steps. Conversely, for the chaotic time series H\'{e}non map and Mackey-Glass, the next time step needs to be predicted. The Perceptual Decision Making and Go/No-go task require the integration of an input signal over a number of time steps in order to solve a classification task afterwards. For the Perceptual Decision Making task the model has to classify which of two presented time series is larger on average, for the Go/No-go task it has to be decided whether or not a "Go" signal has been given. In the appendix in Figure \ref{fig:tasks} the five tasks are depicted. The tasks require different "skills" from the model. For each task, 15 instances are randomly sampled, saved and then loaded during runtime, such that original and null models compute with the same timeseries. 70\% of the data is used to train the readout module. On the remaining 30\% the performance is calculated. The loss functions have been picked in accordance with the style of the task. For the regression tasks of memorizing or predicting a time series, Pearson’s correlation coefficient is used as a loss. For the later statistical analysis, it has been squared. Furthermore, the root mean squared error (RMSE) was used to measure performance. For the classification tasks, the filtered accuracy score is used, which takes into account that there are different segments to these tasks: First, an input signal is presented without a classification required and only afterwards a decision is prompted during a small number of time steps. Only during those later time steps performance is measured.

In order to further examine the results and their statistical significance, a beta regression with fixed effects was fitted to the data. Using the Pearson's correlation coefficient and an accuray metric allows to fit a single model to the whole data. The response variable is the performance measured with the metrics outlined above, but for further analysis the difference between original and null models was used. Additionally, a gaussian or gamma regression was fitted to the results of new runs of the Memory Capacity, H\'{e}non-map and Mackey-Glass prediction tasks measured with the RMSE loss.

In a first pooled analysis the effects incorporated into the model are the type of connectome, the reservoir configuration (series), and age which leads to following formula:
\begin{equation}
    outcome \sim network \times connectome + network \times age + network \times series
\end{equation}
Only two-way terms have been used and it became clear from the results that averaging over all types of connectomes might hide interesting insights. For that reason, the analysis was repeated for each type of connectome individually without the connectome type as a variable:
\begin{equation}
    outcome \sim network \times age + network \times series
\end{equation}
A significance level of 0.05 has been used to identify statistical significance.

\section{Results and Analysis}\label{sec:results}

In this section we will first present the results obtained fitting a beta model to the performance measured with Pearson's correlation coefficient for the regression and filtered accuracy for the classification tasks. Afterwards the performance measured with RMSE on the regression tasks is presented.

\begin{figure}[h]
\centering
\includegraphics[width=0.65\textwidth]{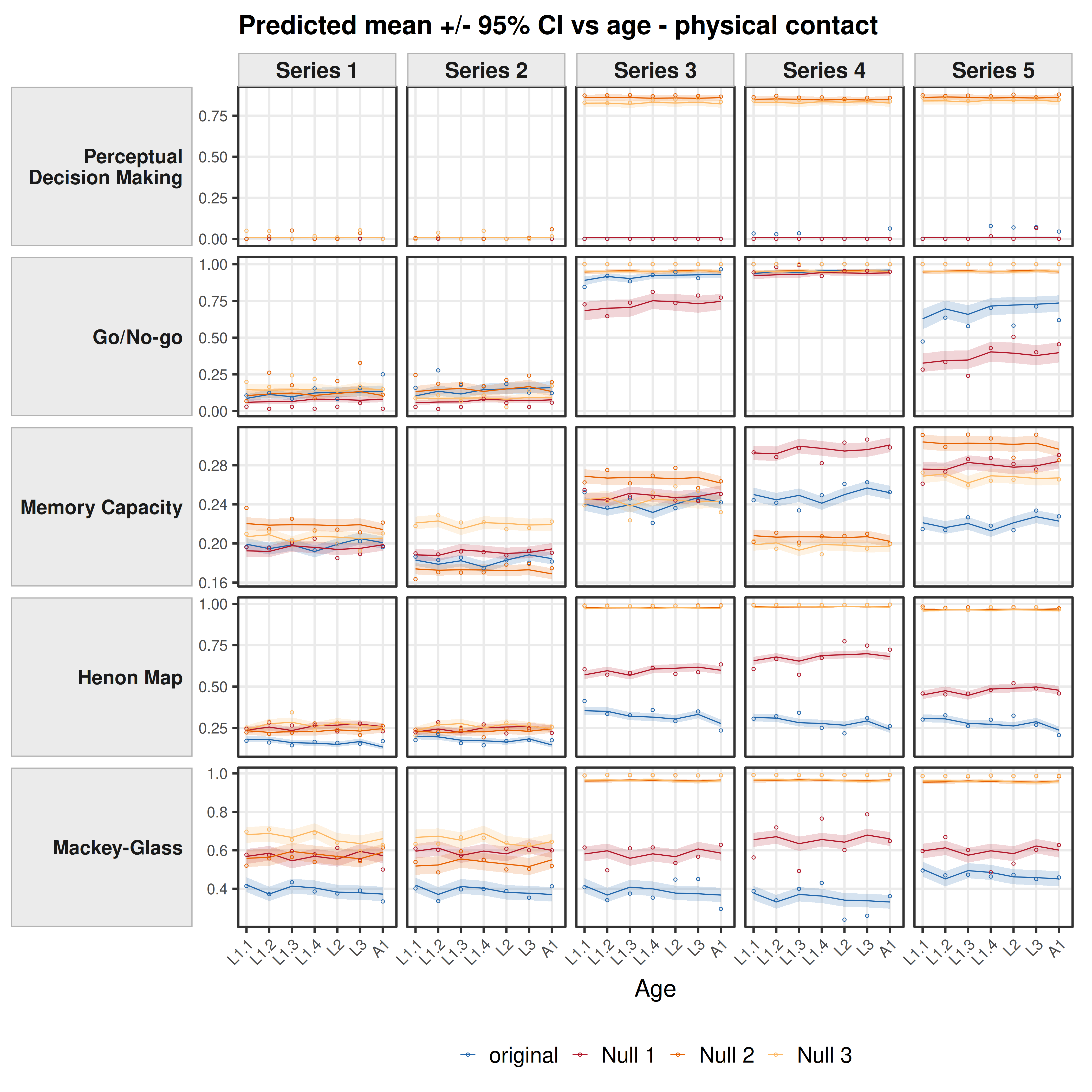}
\caption{The performance of the connectomes based on the number of synapses and the corresponding null models. The rows correspond to the different tasks, the columns to the different reservoir configurations. In the individual plots, the model-based estimated means ± 95\% CI is plotted against the age of the organism.}\label{fig:profile_syn_count}
\end{figure}

\begin{figure}[h]
\centering
\includegraphics[width=0.65\textwidth]{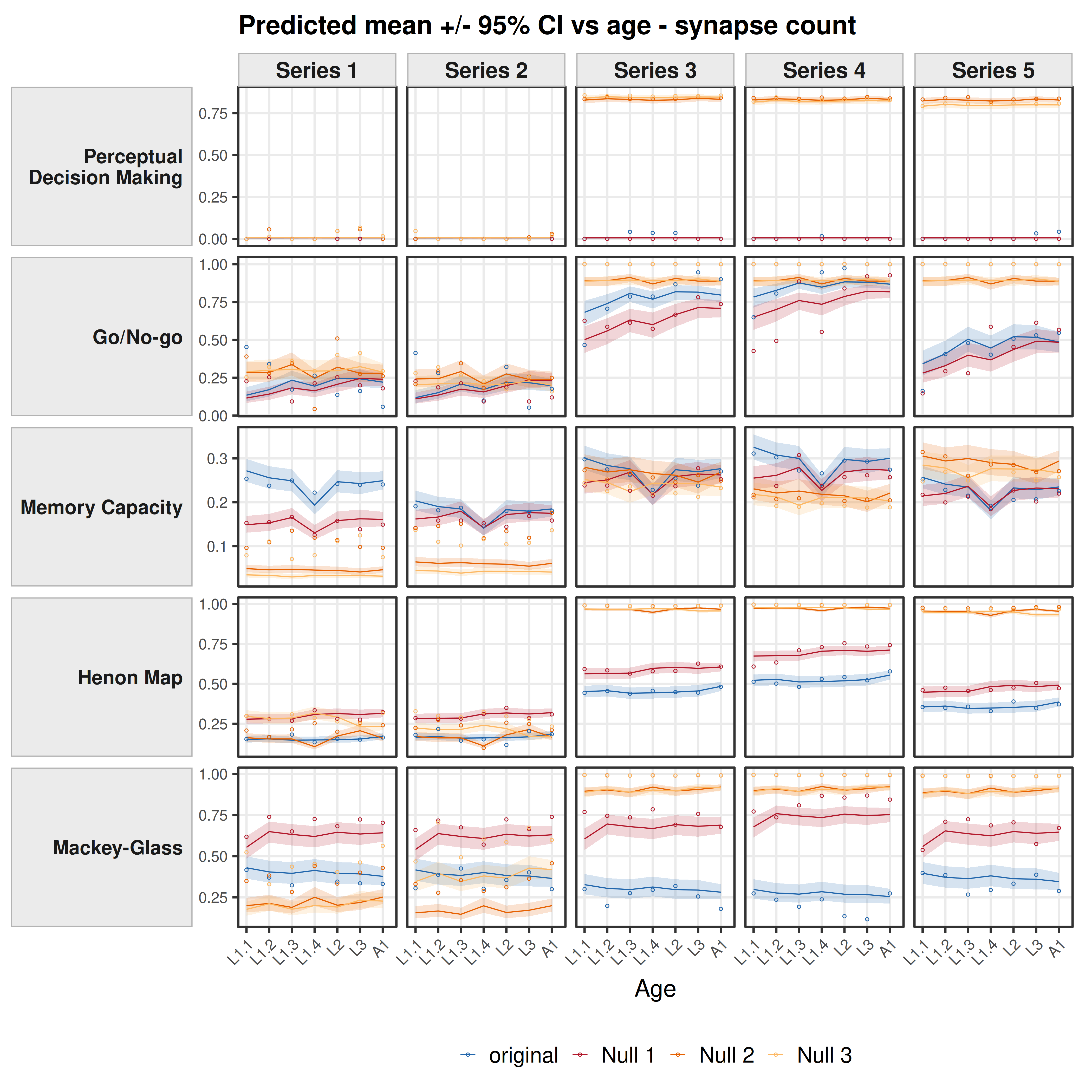}
\caption{The performance of the connectomes based on physical contact and the corresponding null models. The rows correspond to the different tasks, the columns to the different reservoir configurations. In the individual plots, the model-based estimated means ± 95\% CI is plotted against the age of the organism.}\label{fig:profile_phys_contact}
\end{figure}

In Figure \ref{fig:profile_syn_count} and Figure \ref{fig:profile_phys_contact}, the performance of the original and the null models can be seen for the different tasks, reservoir configuration and the method of obtaining the connectome. Upon visual inspection, no clear trends with regards to the age can be seen, meaning that an increase in age does not correspond to an increase or decrease in performance. Furthermore it becomes clear, that in some instances the original or null models have near perfect or near zero performance. Another result, which is also reflected in the results of the statistical analysis, is that the second and third null model often have similar performance and that the connectomes derived from the synaptic size and synaptic number perform quite similarly. For that reason, the following discussion will focus on the first and second null model and a comparison between the connectomes derived from physical contact area on the one side and the synaptic connectomes on the other side. A link to the code repository with further information on all obtained results can be found in the supplementary material.

As can be seen in Table \ref{tab:july-pooled-summary-null123}, the performance often varies significantly for different ages, reservoir configuration and connectome type. Only the Perceptual Decision Making task often does not show significant variation.

For the Memory Capacity task, Tabel \ref{tab:july-pooled-summary-null123} shows variation over age, but also a very strong variation for different types of connectomes. In the appendix, Figure \ref{fig:con_null_1} shows a negative performance impact of the physical contact connectome and a positive impact for the other two connectome types. For this reason, the performance margin for different ages for the Memory Capacity task in Figure \ref{fig:div_null_1} has to be seen critically. What seems like a small performance margin might just be the result of averaging over the different connectome types. For that reason, as mentioned above, the analysis was repeated for each connectome type individually. These findings confirmed the assumption that signs can flip between the results for the physical contact connectome and the two synaptic connectomes. This shows that connectomes derived from the same organism in different ways might show different benchmark results.

\begin{sidewaystable}
\caption{Pooled analysis with beta regression on logit scale. Network row: logit-scale coefficient for original $-$ null (Est., $p$). Omnibus rows: likelihood-ratio $\chi^{2}$ and $p$ for dropping $\mathrm{network}\times\mathrm{Age}$, $\mathrm{network}\times\mathrm{Series}$, or $\mathrm{network}\times\mathrm{connectome}$ from $outcome_{adj} \sim network * connectome + network * Age + network * Series$. Positive network $\Rightarrow$ original higher score.}\label{tab:july-pooled-summary-null123}
\begin{tabular*}{\textheight}{@{\extracolsep\fill}lcccccccccc}
\toprule
Effect & \multicolumn{2}{@{}c@{}}{MC} & \multicolumn{2}{@{}c@{}}{H\'enon} & \multicolumn{2}{@{}c@{}}{MG} & \multicolumn{2}{@{}c@{}}{PDM} & \multicolumn{2}{@{}c@{}}{Go/No-go} \\\cmidrule{2-3}\cmidrule{4-5}\cmidrule{6-7}\cmidrule{8-9}\cmidrule{10-11}
 & Est./$\chi^{2}$ & $p$ & Est./$\chi^{2}$ & $p$ & Est./$\chi^{2}$ & $p$ & Est./$\chi^{2}$ & $p$ & Est./$\chi^{2}$ & $p$ \\
\midrule
\multicolumn{11}{l}{\textbf{Null 1}} \\
$\hat{\beta}$ (network) & $0.305$ & $<.001^{***}$ & $-1.342$ & $<.001^{***}$ & $-0.347$ & $.001^{**}$ & $0.005$ & .965 & $0.282$ & .065 \\
$\mathrm{network}\times\mathrm{Age}$ & $126.87$ & $<.001^{***}$ & $44.52$ & $<.001^{***}$ & $35.39$ & $<.001^{***}$ & $0.20$ & 1.000 & $8.85$ & .182 \\
$\mathrm{network}\times\mathrm{Series}$ & $332.48$ & $<.001^{***}$ & $102.82$ & $<.001^{***}$ & $295.84$ & $<.001^{***}$ & $0.76$ & .944 & $46.62$ & $<.001^{***}$ \\
$\mathrm{network}\times\mathrm{connectome}$ & $534.37$ & $<.001^{***}$ & $257.08$ & $<.001^{***}$ & $220.96$ & $<.001^{***}$ & $0.29$ & .866 & $15.80$ & $<.001^{***}$ \\
\midrule
\multicolumn{11}{l}{\textbf{Null 2}} \\
$\hat{\beta}$ (network) & $0.657$ & $<.001^{***}$ & $-1.086$ & $<.001^{***}$ & $0.009$ & .939 & $-0.098$ & .376 & $-0.386$ & $.010^{*}$ \\
$\mathrm{network}\times\mathrm{Age}$ & $12.69$ & $.048^{*}$ & $2.64$ & .853 & $29.33$ & $<.001^{***}$ & $1.02$ & .985 & $20.06$ & $.003^{**}$ \\
$\mathrm{network}\times\mathrm{Series}$ & $1100.68$ & $<.001^{***}$ & $4091.49$ & $<.001^{***}$ & $2596.95$ & $<.001^{***}$ & $5427.18$ & $<.001^{***}$ & $308.58$ & $<.001^{***}$ \\
$\mathrm{network}\times\mathrm{connectome}$ & $186.90$ & $<.001^{***}$ & $298.18$ & $<.001^{***}$ & $8.44$ & $.015^{*}$ & $9.78$ & $.008^{**}$ & $21.45$ & $<.001^{***}$ \\
\midrule
\multicolumn{11}{l}{\textbf{Null 3}} \\
$\hat{\beta}$ (network) & $0.951$ & $<.001^{***}$ & $-1.426$ & $<.001^{***}$ & $-0.373$ & $.003^{**}$ & $-0.090$ & .421 & $-0.547$ & $<.001^{***}$ \\
$\mathrm{network}\times\mathrm{Age}$ & $28.82$ & $<.001^{***}$ & $5.48$ & .483 & $25.94$ & $<.001^{***}$ & $0.31$ & .999 & $21.08$ & $.002^{**}$ \\
$\mathrm{network}\times\mathrm{Series}$ & $1039.60$ & $<.001^{***}$ & $3577.39$ & $<.001^{***}$ & $1897.27$ & $<.001^{***}$ & $5018.84$ & $<.001^{***}$ & $315.41$ & $<.001^{***}$ \\
$\mathrm{network}\times\mathrm{connectome}$ & $223.17$ & $<.001^{***}$ & $232.26$ & $<.001^{***}$ & $8.30$ & $.016^{*}$ & $1.67$ & .433 & $31.20$ & $<.001^{***}$ \\
\botrule
\end{tabular*}
\footnotetext{Note: Reference: network = null. Significance: $*$ $p<.05$, $**$ $p<.01$, $***$ $p<.001$.}
\end{sidewaystable}

When comparing the original model with the first null model, which has a random wiring but similar I/O nodes, it can be seen that for the Perceptual Decision Making task, there is often little difference between the original and the null models. For the conceptually similar Go/No-go task, the original models will often perform better. For the Memory Capacity task, this depends on the choice of the connectome, as written above. For the two prediction tasks, there is a clear advantage of the null models. Thus, the biological wiring seems like it can under certain circumstances present a slight advantage for tasks requiring the integration of data, while being a disadvantage for the prediction of timeseries.

\begin{landscape}
\begin{figure}[p]
\centering
\begin{minipage}[t]{0.48\textwidth}
    \centering
    \includegraphics[height=0.8\textheight]{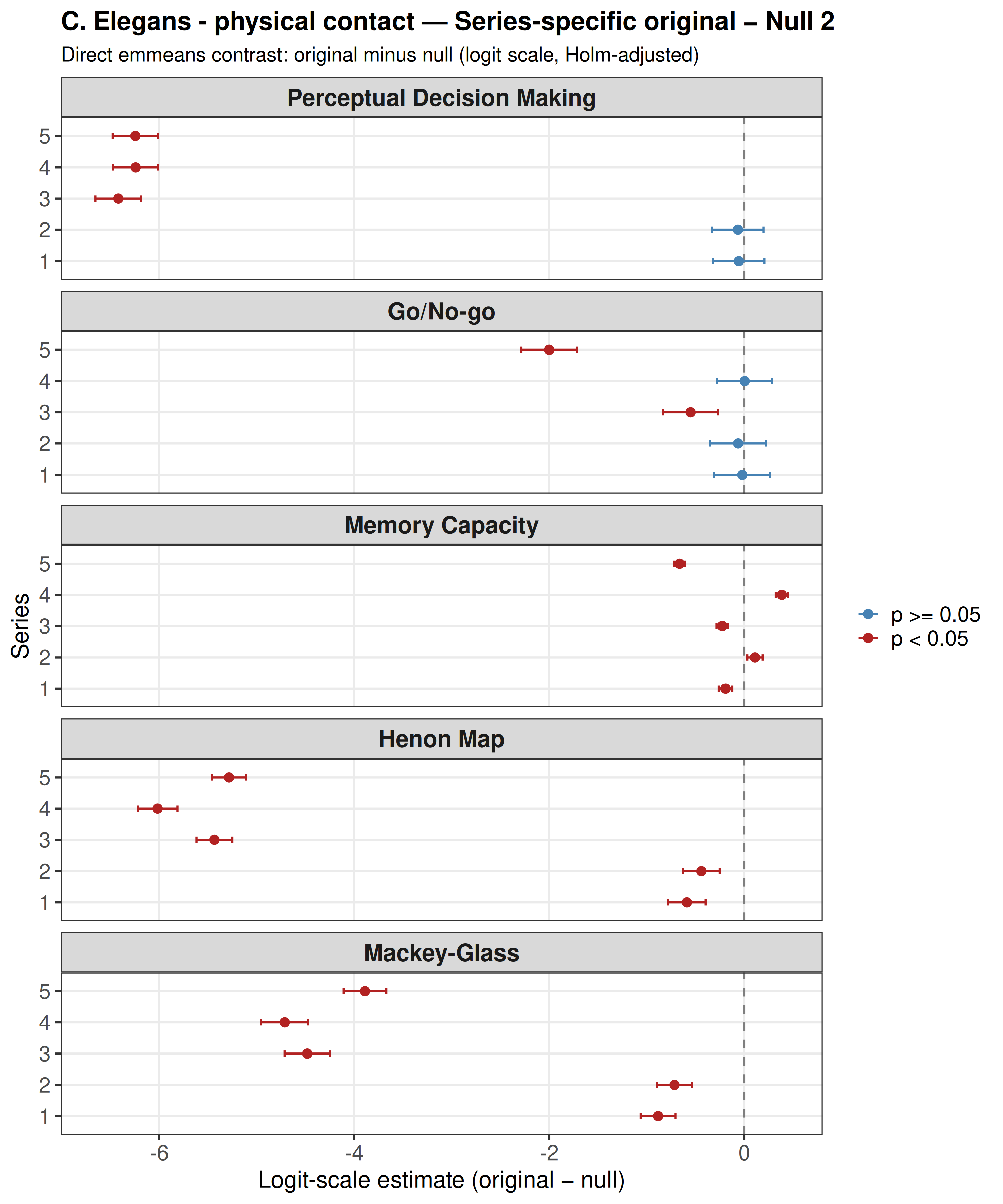}
    \caption{The series-specific simple contrasts for the five different tasks from the stratified analysis for the connectome derived from physical contact and the second null model. For the five different reservoir configurations, the estimated marginal $original-null_2$ difference on the logit scale with Holm-adjustment within each task is depicted. The bars mark confidence intervals, the color codes statistical significance.}
    \label{fig:strat_phys_ser_null_2}
\end{minipage}\hspace{0.25\textwidth}
\begin{minipage}[t]{0.48\textwidth}
    \centering
    \includegraphics[height=0.8\textheight]{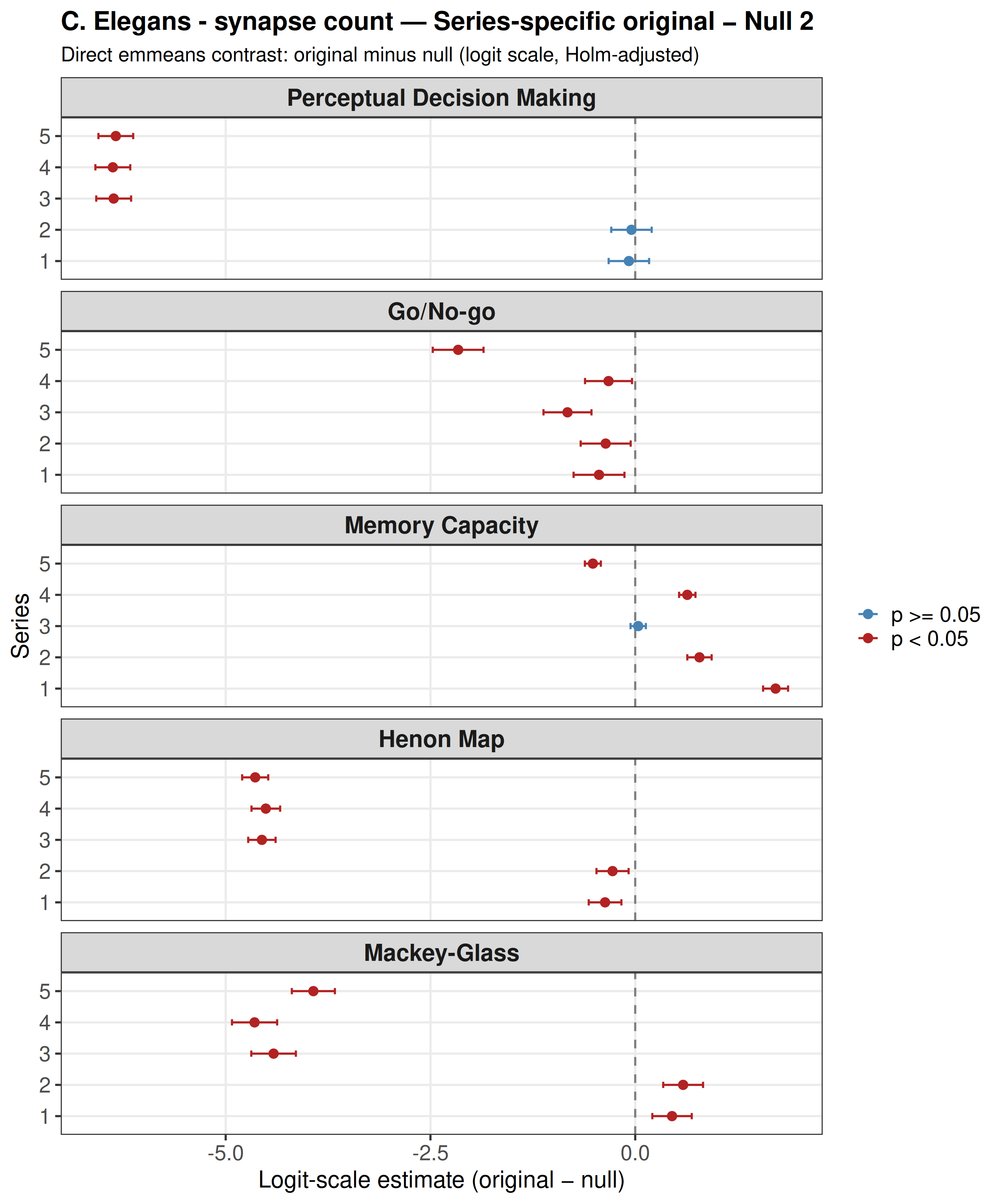}
    \caption{The series-specific simple contrasts for the five different tasks from the stratified analysis for the connectome derived from the number of connecting synapses and the second null model. For the five different reservoir configurations, the estimated marginal $original-null_2$ difference on the logit scale with Holm-adjustment within each task is depicted. The bars mark confidence intervals, the color codes statistical significance.}
    \label{fig:strat_count_ser_null_2}
\end{minipage}
\end{figure}
\end{landscape}

The other two null models, which have randomized I/O nodes, outperform the original model especially for the reservoir configurations that allow a high number of I/O nodes to be picked, as can be seen for the Perceptual Decision Making task and the H\'{e}non map and Mackey-Glass prediction in Figure \ref{fig:strat_phys_ser_null_2}. Comparing this to Figure \ref{fig:strat_count_ser_null_2}, the connectome type might shift results without changing the overall pattern, which leads to the synaptically derived connectomes outperforming the null models on the Memory Capacity task.
Considering how the results depend on the reservoir configuration, one might argue that a biological I/O node choice is beneficial in cases in which the total number of I/O nodes is restricted. The performance of the second null model, which only has randomized I/O nodes, and the third null model, which also randomizes the wiring, are similar. Both together differ from the performance of the first null model. Thus it seems like randomizing the I/O nodes has a larger impact than the rewiring of the network itself.

When using the RMSE as a performance measure the results largely agree for the H\'{e}non map and Mackey-Glass prediction. For the Memory Capacity task, the contrasts for different ages and series show some different results when measured with RMSE than when measured with Pearson's correlation coefficient. The synaptic connectomes do not show a better performance than the Null models anymore, although it has to be noted that the contrasts where relatively small for that task. This can be seen in Table \ref{tab:mc-null1-age-csv}, where the mean age contrasts flips for the synapse based connectomes, resulting in the null models having favourable performance.

\begin{table}[h]
\caption{Simple contrasts (original $-$ $Null_1$) for age for the Memory Capacity Task. Beta regression with Pearson's correlation coefficient scores: positive $\Rightarrow$ original better. Gamma/Gaussian regression with RMSE scores: positive $\Rightarrow$ original worse.}\label{tab:mc-null1-age-csv}
\begin{tabular*}{\textwidth}{@{\extracolsep\fill}lcccccc}
\toprule
& \multicolumn{3}{@{}c@{}}{Pearson's correlation coefficient} & \multicolumn{3}{@{}c@{}}{RMSE} \\\cmidrule{2-4}\cmidrule{5-7}
Connectome & Mean Age & Fav.\ orig. & Fav.\ null & Mean Age & Fav.\ orig. & Fav.\ null \\
\midrule
phys\_contact  & $-0.20$ & $0$ & $7$ & $+0.044$ & $0$ & $7$\\
syn\_count     & $+0.35$ & $7$ & $0$ & $+0.0065$ & $0$ & $4$\footnotemark[1]\\
syn\_size      & $+0.24$ & $5$ & $0$\footnotemark[2] & $+0.027$ & $0$ & $5$\footnotemark[2]\\
\botrule
\end{tabular*}
\footnotetext{Note: Counts are Holm-adjusted Age levels with $p<0.05$.}
\footnotetext[1]{Three Age levels non-significant.}
\footnotetext[2]{Two Age levels non-significant.}
\end{table}

\section{Conclusion}\label{sec:conclusion}

In this paper, connectomes of C. elegans derived with different measurements and at different ages have been implemented as reservoirs with different I/O configurations and tested on five different neuro-inspired tasks. As a comparison, three different null models have been used to check the impact of biologically-informed wiring and I/O choices.

The differences between the three types of connectomes as described in \cite{witvliet2021connectomes} is reflected in the performance results, showing that results are not random.

Aging has neither been found to consistently increase or decrease the performance of the networks, nor is there a similar pattern of performances over age for the different tasks or series. An open question is to test how perturbations of the network, in the form of deleted nodes or edges, would impact performance, as that had a noticeable effect on the mobile communication networks examined in \cite{reimers2025benchmarking}.

Whether biologically-informed wiring and I/O nodes are an advantage or disadvantage depends on the type of connectome, reservoir configuration, used task and performance measure. In most cases, the null models outperform the original models. This is especially true for random I/O nodes. Further inspection and experiments could be conducted to answer follow-up questions to the research questions of this paper: How can it be explained that the synaptic networks are stronger than the rewired networks on the Memory Capacity task, when in most cases the null models outperform the original models? Is it true that the benefit of a biologically-informed choice of input and output nodes is dependent on a low budget to choose nodes as the results suggest? The answers to these questions might be of interest to the development of neuro-inspired systems as they seem to mark certain constraints and conditions.

\backmatter

\bmhead{Supplementary information}\label{sec:supplement}

The code used for this paper and additional results can be found \href{https://github.com/Deskt0r/celegans_reservoir_benchmark}{here}.

\newpage
\begin{appendix}

\section*{Appendix}

\begin{figure}[h!]
\centering
\begin{minipage}[t]{0.49\textwidth}
    \centering
    \includegraphics[width=\linewidth]{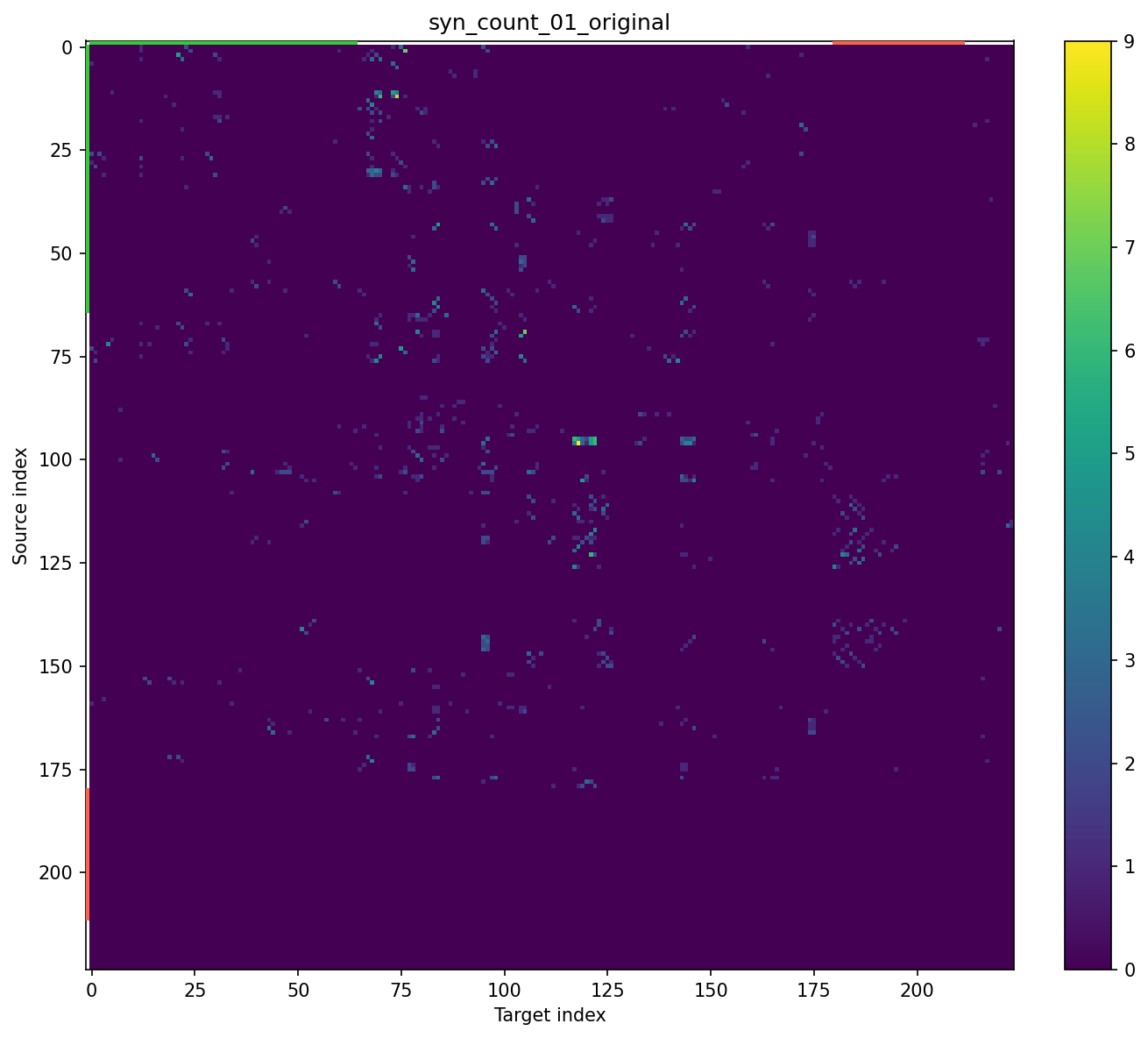}
    \caption{Heatmap of the network derived through counting of synapses from a C. elegans organism at the earliest larvae state (L1.1).}
    \label{fig:heatmap_orig}
\end{minipage}\hspace{0.01\textwidth}
\begin{minipage}[t]{0.49\textwidth}
    \centering
    \includegraphics[width=\linewidth]{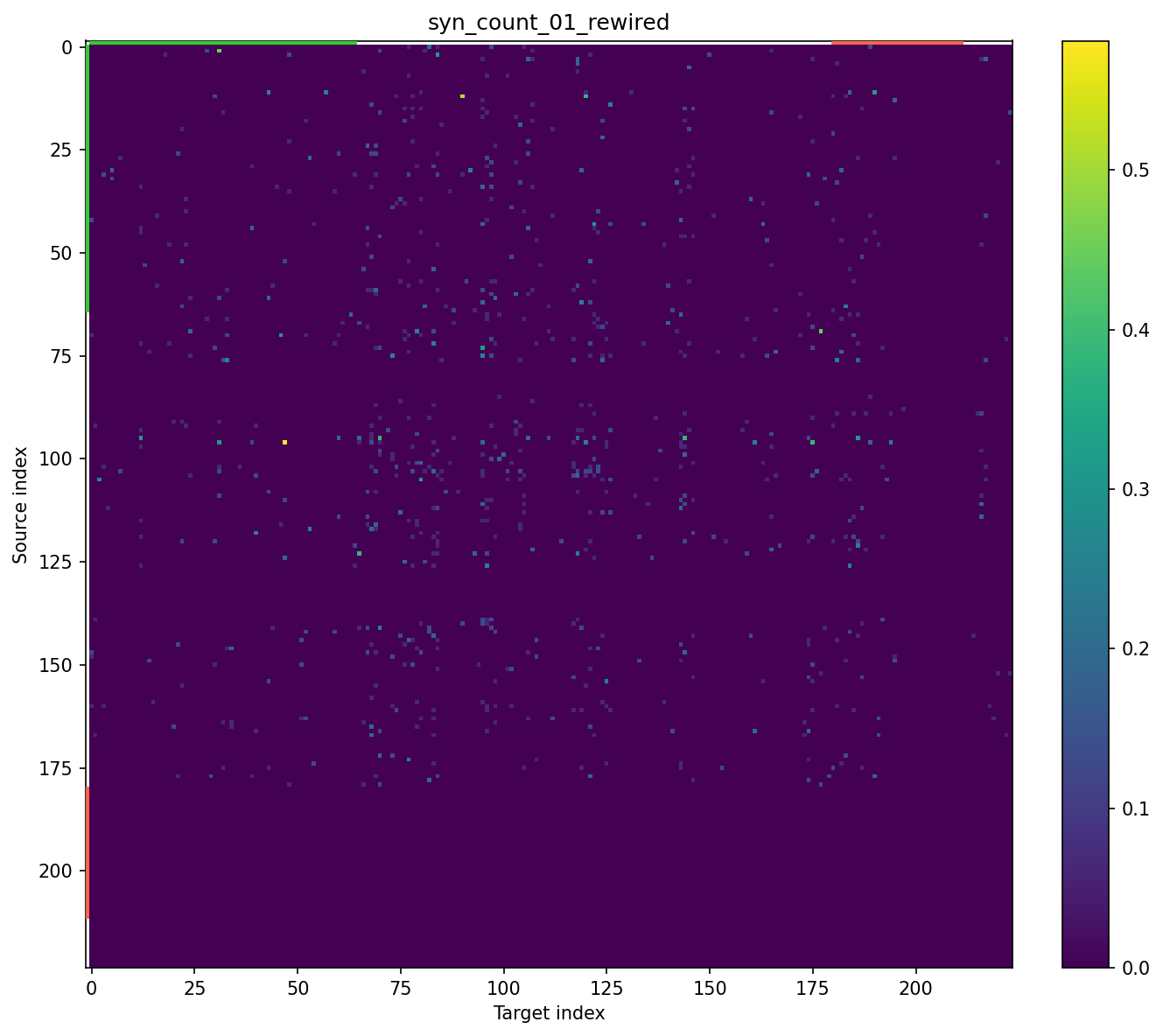}
    \caption{Heatmap of the network derived through counting of synapses from a C. elegans organism at the earliest larvae state (L1.1) after rewiring.}
    \label{fig:heatmap_rewired}
\end{minipage}
\end{figure}

\begin{figure}[h]
\centering
\includegraphics[width=0.8\textwidth]{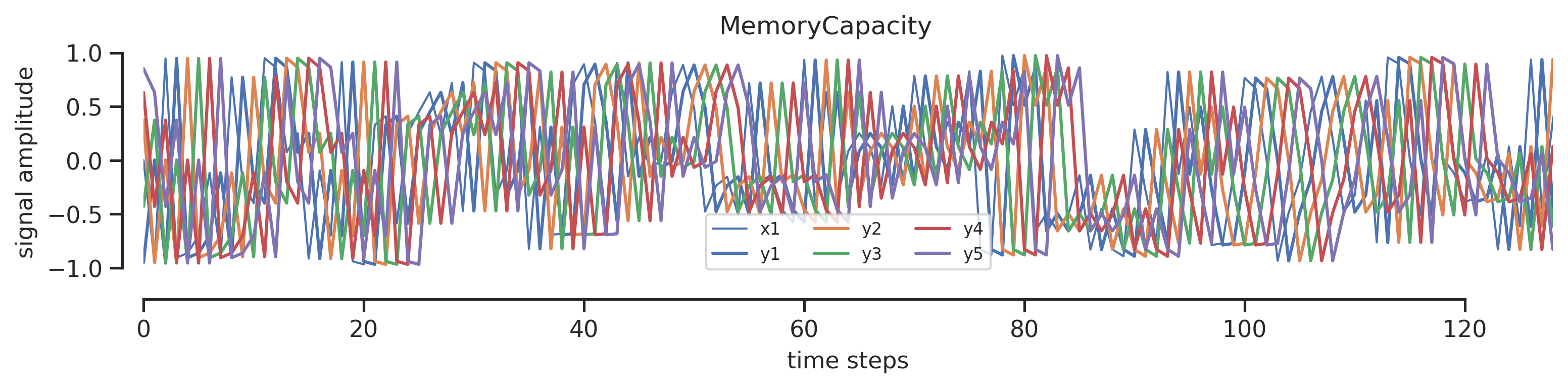}

\vspace{1.5mm}
{\small (a) Time series of the Memory Capacity task}

\vspace{2mm}
\includegraphics[width=0.8\textwidth]{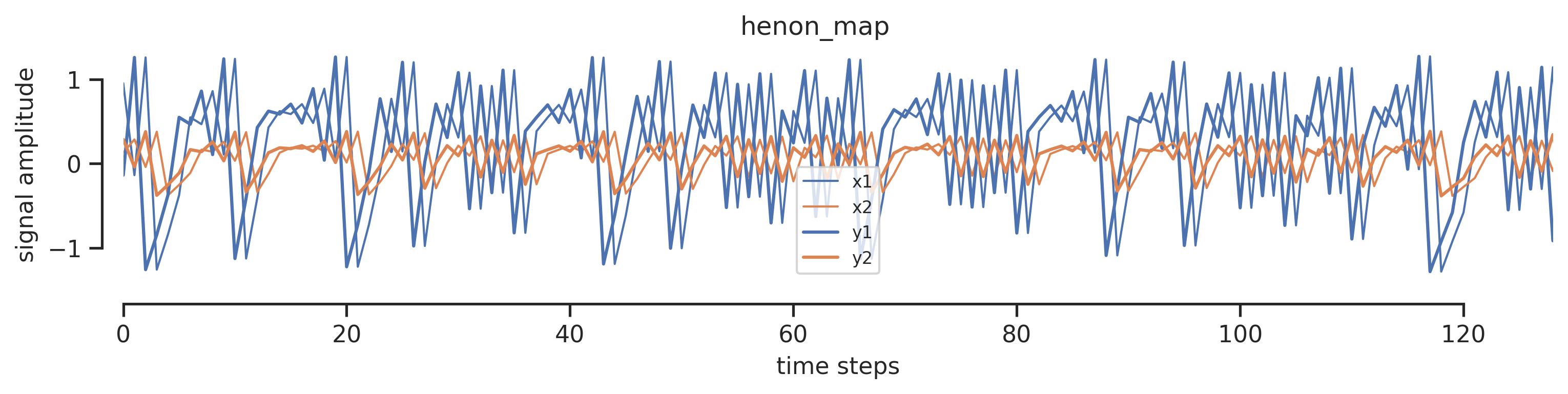}

\vspace{1.5mm}
{\small (b) Time series of the H\'{e}non map prediction task}

\vspace{2mm}
\includegraphics[width=0.8\textwidth]{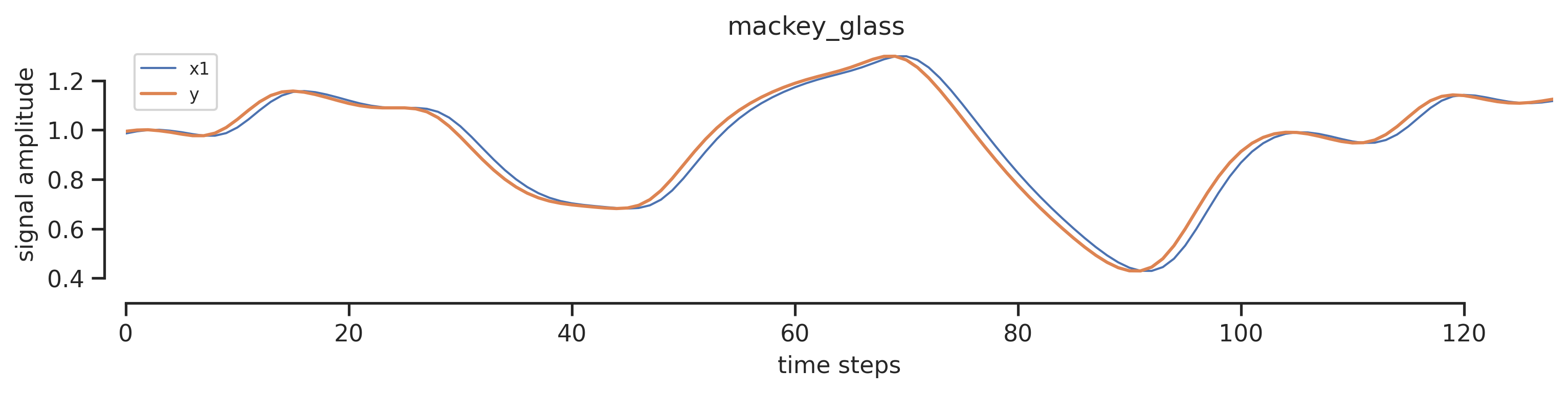}

\vspace{1.5mm}
{\small (c) Time series of the Mackey-Glass prediction task}

\vspace{2mm}
\includegraphics[width=0.8\textwidth]{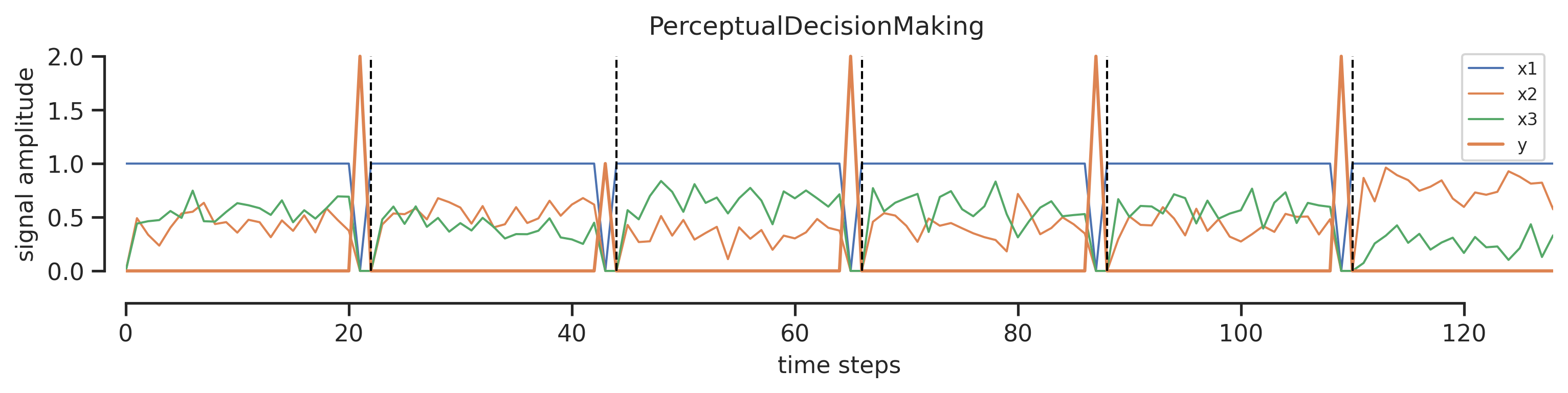}

\vspace{1.5mm}
{\small (d) Time series of the Perceptional Decision Making task}

\vspace{2mm}
\includegraphics[width=0.8\textwidth]{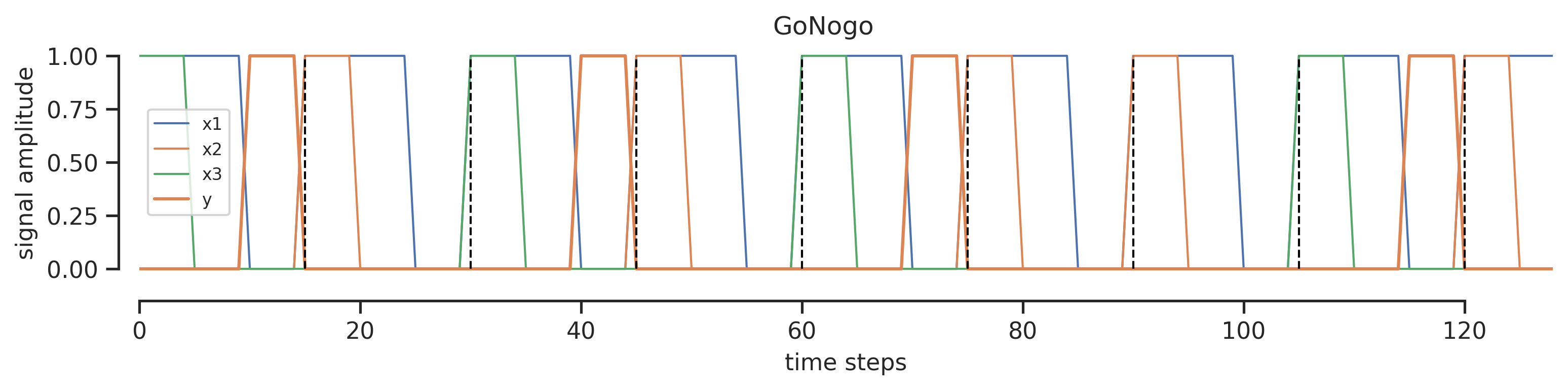}

\vspace{1.5mm}
{\small (e) Time series of the Go/No-go task}

\caption{Plots of the time series data associated with each task.}
\label{fig:tasks}
\end{figure}

\begin{landscape}
\begin{figure}[p]
\centering
\begin{minipage}[t]{0.48\textwidth}
    \centering
    \includegraphics[height=0.8\textheight]{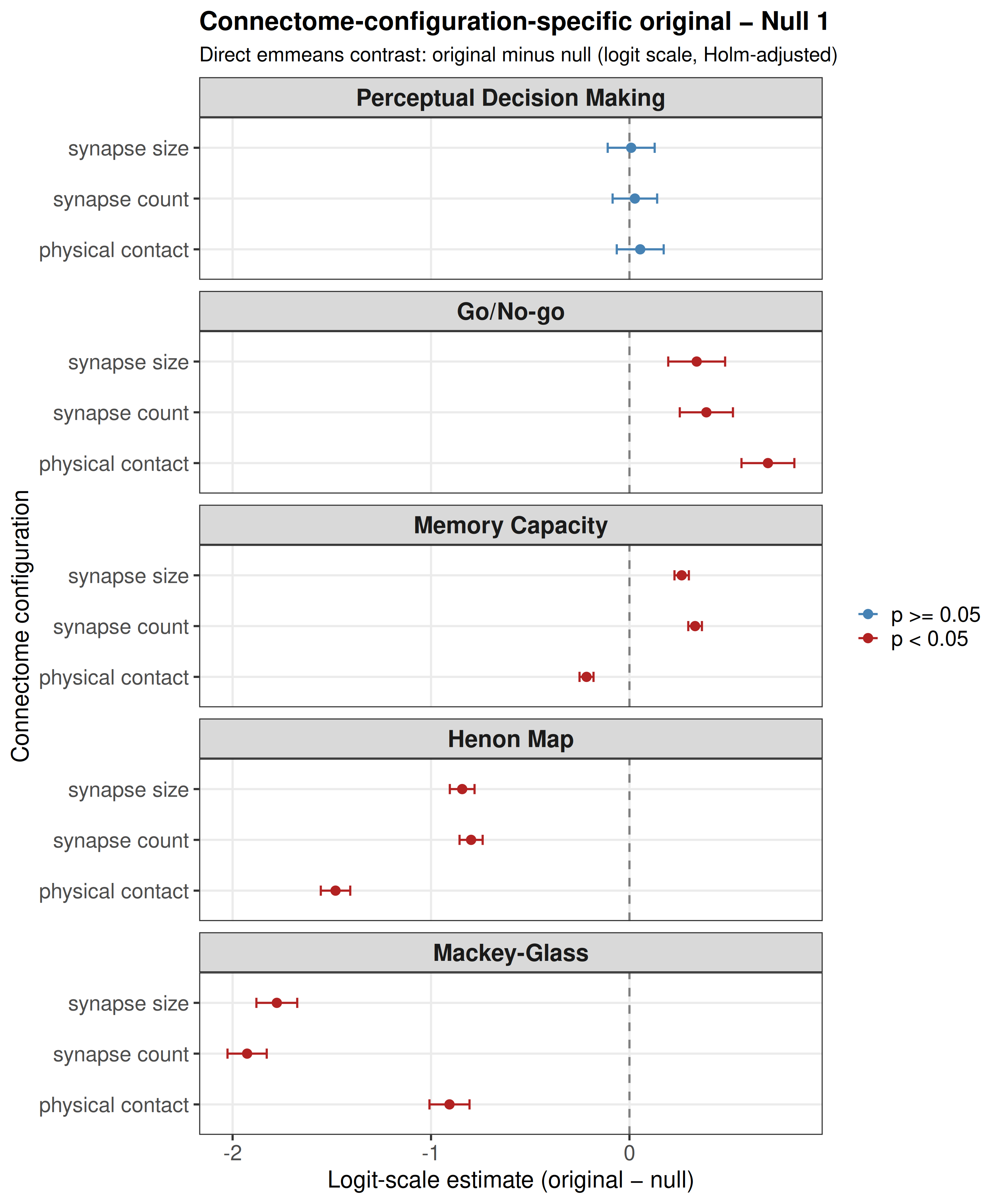}
    \caption{The connectome-specific simple contrasts for the five different tasks from the pooled analysis for the first null model. For the three different types of connectomes (physical contact, synapse count, synapse size), the estimated marginal $original-null_1$ difference on the logit scale with Holm-adjustment within each task is depicted. The bars mark confidence intervals, the color codes statistical significance.}
    \label{fig:con_null_1}
\end{minipage}\hspace{0.25\textwidth}
\begin{minipage}[t]{0.48\textwidth}
    \centering
    \includegraphics[height=0.8\textheight]{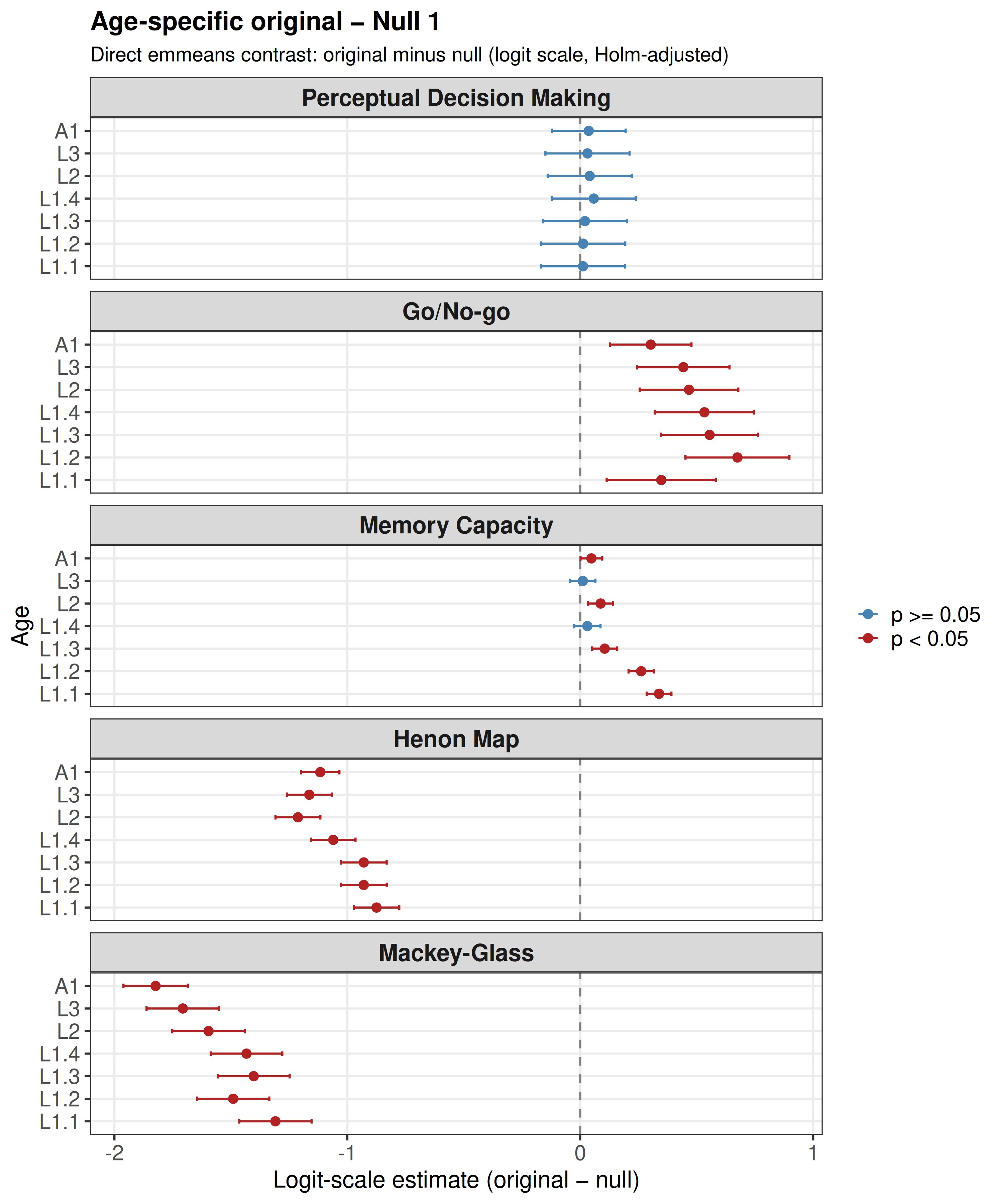}
    \caption{The age-specific simple contrasts for the five different tasks from the pooled analysis for the first null model. For the different ages the estimated marginal $original-null_1$ difference on the logit scale with Holm-adjustment within each task is depicted; the two adult states have been merged prior. The bars mark confidence intervals, the color codes statistical significance.}
    \label{fig:div_null_1}
\end{minipage}
\end{figure}
\end{landscape}

\end{appendix}

\newpage
\bibliography{bibliography_conn}

\end{document}